\documentclass{article}

\usepackage{PRIMEarxiv}

\usepackage[utf8]{inputenc} 
\usepackage[T1]{fontenc}    
\usepackage{hyperref}       
\usepackage{url}            
\usepackage{booktabs}       
\usepackage{amsfonts}       
\usepackage{nicefrac}       
\usepackage{microtype}      
\usepackage{lipsum}
\usepackage{fancyhdr}       
\usepackage{graphicx}       
\graphicspath{{media/}}     
\usepackage{cite}
\usepackage{booktabs,multirow,makecell,tabularx}
\usepackage{amsmath,amssymb,amsfonts}
\usepackage{algorithmic}
\usepackage{graphicx}
\usepackage{textcomp}
\usepackage{xcolor}
\usepackage[hypcap=false]{caption}
\usepackage[none]{hyphenat}
\title{Inference-time Stochastic Refinement of
GRU-Normalizing Flow for\\
Real-time Video Motion Transfer
}

\author{
  Tasmiah Haque \\
  Department of Industrial and Management Systems Engineering \\
  West Virginia University \\
  Morgantown, WV 26505\\
  \texttt{th00027@mix.wvu.edu} \\
   \And
  Srinjoy Das \thanks{Corresponding author} \\
  School of Mathematical and Data Sciences \\
  West Virginia University \\
  Morgantown, WV 26505\\
  \texttt{srinjoy.das@mail.wvu.edu} \\
}

\begin{document}
\maketitle

\begingroup
\renewcommand\thefootnote{}%
\footnotetext{This manuscript is accepted for publication in the Proceedings of the Asilomar Conference on Signals, Systems, and Computers (Asilomar 2025).}
\addtocounter{footnote}{-1}%
\endgroup

\begin{abstract}
Real-time video motion transfer applications such as immersive gaming and vision-based anomaly detection require accurate yet diverse future predictions to support realistic synthesis and robust downstream
decision making under uncertainty. To improve the diversity of such sequential forecasts we propose
a novel inference-time refinement technique that combines Gated Recurrent Unit-Normalizing Flows
(GRU-NF) with stochastic sampling methods. While GRU-NF can capture multimodal distributions
through its integration of normalizing flows within a temporal forecasting framework, its deterministic
transformation structure can limit expressivity. To address this, inspired by Stochastic Normalizing Flows
(SNF), we introduce Markov Chain Monte Carlo (MCMC) steps during GRU-NF inference, enabling
the model to explore a richer output space and better approximate the true data distribution without
retraining. We validate our approach in a keypoint-based video motion transfer pipeline, where capturing temporally coherent and perceptually diverse future trajectories is essential for realistic samples and
low bandwidth communication. Experiments show that our inference framework, Gated Recurrent Unit-
Stochastic Normalizing Flows (GRU-SNF) outperforms GRU-NF in generating diverse outputs without
sacrificing accuracy, even under longer prediction horizons. By injecting stochasticity during inference,
our approach captures multimodal behavior more effectively. These results highlight the potential of
integrating stochastic dynamics with flow-based sequence models for generative time series forecasting. The code is available at: https://github.com/Tasmiah1408028/Inference-Time-Stochastic-Refinement-Of- GRU-NF-For-Real-Time-Video-Motion-Transfer
\end{abstract}


\section{Introduction}
Motion transfer frameworks \cite{b1, b2, b3, b4} operate by extracting low-dimensional keypoints or landmark representations from a driving video and transferring the resulting motion to a static source image to synthesize new frames. Since these keypoints compactly encode pose and deformation, they enable low bandwidth motion representations and minimize the reliance on full-frame video
during transmission, which is attractive for real-time systems. However, many deployed applications require diverse predictions of the near future rather than a single best forecast. The reason is fundamental: human motion and many physical interactions are multimodal and therefore the same current pose can evolve into multiple, equally plausible futures.

\begin{figure}
\centering
    \includegraphics[width=0.5\textwidth,height=0.25\textheight]{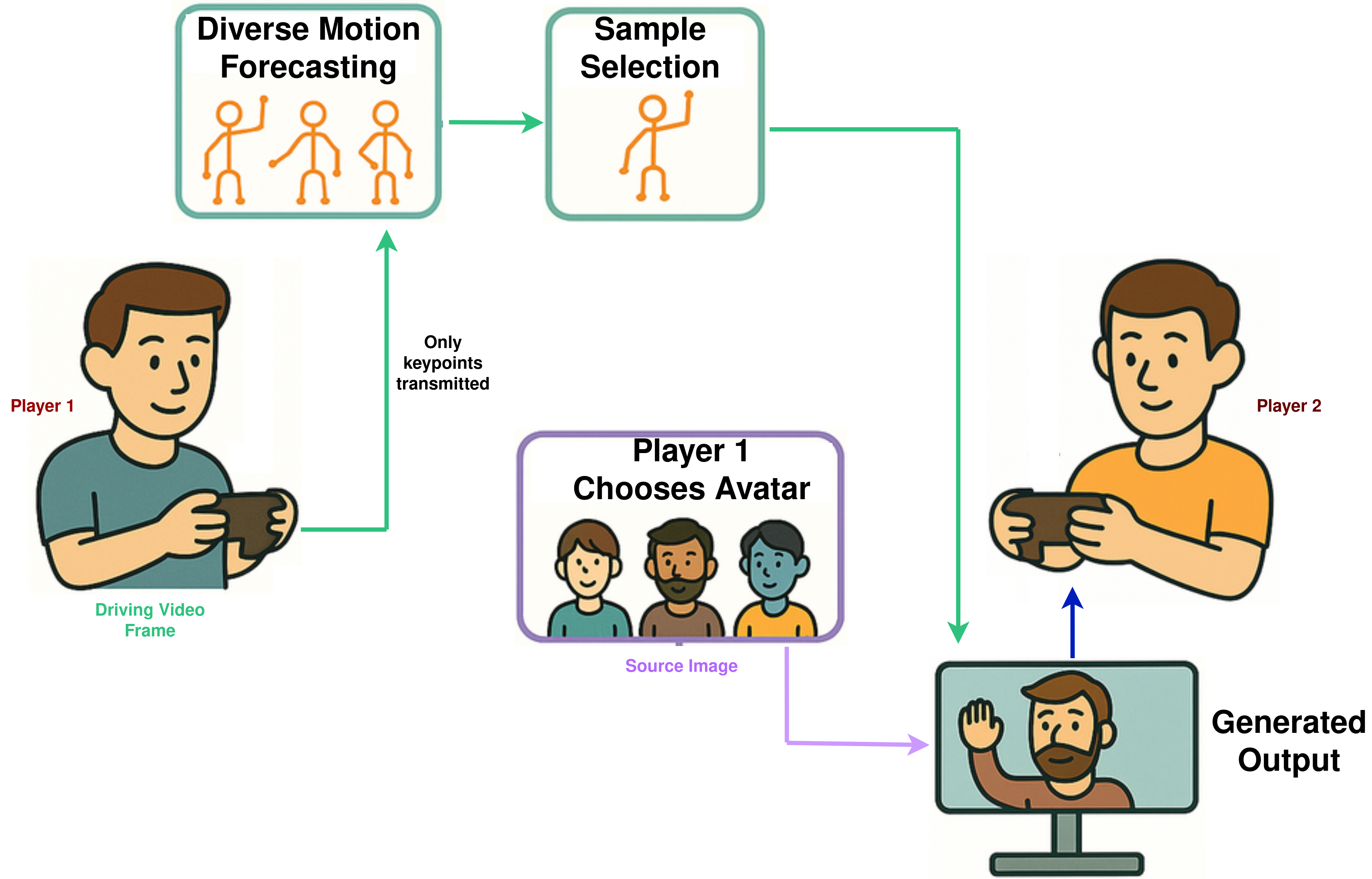}
    \caption{Diverse sample prediction in real-time virtual reality (VR) gaming}
    \label{fig:VR gaming}
\end{figure}

\begin{figure*}[t]
\centering
    \includegraphics[width=0.85\textwidth,height=0.35\textheight]{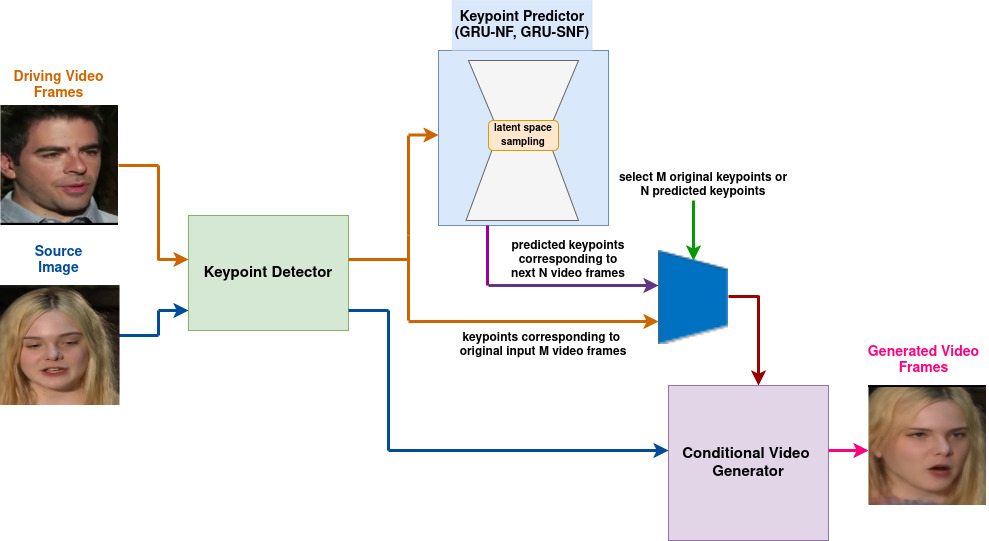}
    \caption{The proposed pipeline for real-time video motion transfer for bandwidth reduction.}
    \label{fig:proposed pipeline}
\end{figure*}

Two representative use cases illustrate this need for diversity. In virtual reality (VR) gaming, users perform unscripted actions; identical poses (e.g., a neutral stance) may precede different motions as shown in Figure~\ref{fig:VR gaming}. To preserve immersion and responsiveness, a motion-transfer model should produce multiple coherent samples that cover these alternatives while remaining temporally smooth. In manufacturing anomaly detection, visual inspection systems often face limited labeled defect data and class imbalance. Forecasting multiple plausible futures of the scene can expose subtle, low-frequency anomalies that a single predicted trajectory would miss; these samples can also act as targeted data augmentation to improve the robustness of anomaly classifiers trained on top of the generated frames. For such real-time deployments, motion-transfer pipelines benefit from an explicit prediction module (Figure~\ref{fig:proposed pipeline}) that forecasts future keypoints and feeds them to the renderer, achieving bandwidth reduction while still enabling synthesis of diverse high quality video frames. In this framework keypoints are first extracted from source and driving videos following which multiple plausible future keypoint trajectories are forecast from a single input sequence of keypoints corresponding to the driving video using the latent space of the generative time series model. This maintains bandwidth efficiency both spatially (keypoints are transmitted instead of video frames) and temporally (for M+N frames only the first M are transmitted and the next N are generated using prediction). The keypoints are then synthesized into corresponding future video frames through the conditional video generator.

Existing probabilistic sequence models 
such as Variational Recurrent Neural Networks (VRNNs) \cite{b5} augment recurrent predictors with stochastic latent variables trained via variational inference in the Variational Autoencoder (VAE) framework \cite{b28}, and can provide strong one-step or short-horizon predictive accuracy on time series whose future evolution is inherently stochastic or multimodal. However, because training relies on an approximate variational objective, VRNNs may under-utilize the latent space or collapse modes, which can reduce the diversity of sampled futures in practice \cite{b6}. On the other hand, Gated Recurrent Unit–Normalizing Flow (GRU-NF) models \cite{b7} combine powerful temporal representations (GRUs) with exact-likelihood based normalizing flows \cite{b8}. However, the invertibility constraint of such flow based networks imposes a structural bias that can make it difficult to map between distributions with well-separated modes, a limitation that is often described as the topological constraint of flows \cite{b9}. In practice, this can manifest as mode under-coverage 
especially at long horizons where compounding uncertainty is substantial.

To address these gaps, we propose Gated Recurrent Unit–Stochastic Normalizing Flow (GRU-SNF), inspired by Stochastic Normalizing Flows (SNF) \cite{b9}. Unlike SNF, which incorporates stochastic refinement during training, GRU-SNF keeps a trained GRU-NF fixed and inserts a small number of Markov Chain Monte Carlo (MCMC) refinement steps \cite{b10} at inference. Specifically, each flow sample initializes a short MCMC chain that proposes local perturbations and accepts or rejects them using an energy that balances prior plausibility (staying on the learned manifold) with temporal consistency. This post-hoc refinement nudges samples toward higher-probability regions to improve mode coverage without retraining, and a few steps per timestep typically suffice while preserving real-time latency.

\section{Related Work}

\begin{itemize}
    \item \textbf{Diverse video sample prediction:}
A major class of diverse video predictors extends deterministic recurrent models with stochastic latent variables so the same past can yield multiple plausible futures. Video-specific examples include SV2P \cite{b12} and SVG-LP \cite{b13}, which learn priors over latents to sample multiple rollouts, and SAVP \cite{b14}, which combines a VAE objective with adversarial training to encourage realism. These methods often face tradeoffs such as blurry frames from likelihood based training, instability in training with adversarial losses, and long-horizon drift from compounding errors, especially while generating in pixel-space. Diffusion-based generators (e.g., Video Diffusion Models, Masked Conditional Video Diffusion, Latent Video Diffusion) improve quality and diversity via iterative denoising \cite{b16,b17,b18}, but typically require many sampling steps and large backbones, making them expensive for real-time or low-power deployment. Maintaining temporal consistency over long videos also remains challenging \cite{b19}. Flow-based video models such as VideoFlow \cite{b15} offer exact likelihoods and explicit temporal structure,
however invertibility constraints and full-frame modeling can still be computationally heavy. Because many of these approaches generate full video frames, they are compute and bandwidth-intensive. In contrast, keypoint-based motion transfer schemes (e.g. FOMM) use low-dimensional keypoints for generating frames. In such frameworks GRU-NF enables diverse keypoint trajectory prediction with coherent rendered video samples \cite{b6}. Our work keeps this efficient keypoint-space setup and improves diversity through lightweight inference-time sampling refinements on top of a trained GRU-NF.
\end{itemize}

\begin{itemize}
    \item \textbf{Inference-time refinement for diverse video generation:} A complementary line of work improves samples after or during generation without retraining the base model, for various applications including video generation. Video diffusion models widely employ inference-time guidance e.g., classifier guidance \cite{b20} and classifier-free guidance \cite{b21} to steer samples toward higher-utility regions without modifying training.  In energy-based modeling, samples are often obtained via Langevin dynamics / MCMC, i.e., noisy gradient-based moves on an energy landscape, and this has been used effectively for sequence and video generation \cite{b22,b23}. Our approach adds a few lightweight MCMC refinements at inference, guided by an energy function that mixes prior plausibility with GRU-based predictive consistency, yielding better mode coverage with minimal inference-time overhead. Conceptually, GRU-SNF differs from other inference-time refinement schemes such as Discriminator Rejection Sampling (DRS) \cite{b11}, which filter or reweight already generated samples. Rather than discarding samples, our approach reshapes them through lightweight, principled stochastic moves guided by a task-aware energy function. This distinction matters for interactive systems: refinement within the sampling process yields higher acceptance of diverse yet coherent futures with minimal computational overhead, whereas external rejection can waste compute and reduce throughput.
\end{itemize}

\section{Proposed Methodology}

Our baseline model GRU-NF combines the sequential modeling ability of GRU with NF which learns a differentiable, invertible mapping $f$ between real-world data \(\boldsymbol{x}\) and a latent variable \(\boldsymbol{z}\) with a tractable prior \(p(\boldsymbol{z})\). 
The density of \(\boldsymbol{x}\) is given as follows: 

\begin{equation}
\label{eq:NF}
p_X(\boldsymbol{x}) = p_Z(f(\boldsymbol{x})) \left| \det \left( \frac{\partial f(\boldsymbol{x})}{\partial \boldsymbol{x}^T} \right) \right|
\end{equation}

\noindent where \( \frac{\partial f(\boldsymbol{x})}{\partial \boldsymbol{x}^T} \) is the Jacobian of $f$ at \(\boldsymbol{x}\).

\noindent During inference, the GRU component of the GRU-NF captures temporal dependence by using the past input \(\boldsymbol{y}_{t-1}\) and hidden state \(\boldsymbol{h}_{t-1}\) corresponding to the previous timestep as follows:

\begin{equation}
\boldsymbol{h}_{t} = \text{GRU}(\boldsymbol{y}_{t-1}; \boldsymbol{h}_{t - 1})
\end{equation} 

\noindent The predicted value \(\boldsymbol{y}_t\) at time $t$ is then generated using \(\boldsymbol{z}_t \sim N(0,I)\) and the GRU hidden state \(\boldsymbol{h}_t\) which captures the sequence history:


\begin{equation}
{\boldsymbol{y}}_{t} = f^{-1}(\boldsymbol{z}_{t} \mid {\boldsymbol{h}}_{t})
\end{equation}

\noindent In our proposal the above described inference steps of the GRU-NF are extended with MCMC sampling  to effectively capture multimodal target distributions. Between each normalizing flow transformation  \(\boldsymbol{y}_{t,k} \rightarrow \boldsymbol{y}_{t,k+1}\) at layer $k$, we apply a stochastic sampling step governed by a potential function \(u_\lambda(\boldsymbol{y})\) \cite{b9}:

\begin{equation}
u_\lambda(\boldsymbol{y}) = (1 - \lambda) u_{\boldsymbol{Z}}(\boldsymbol{y}) + \lambda u_{\boldsymbol{X}}(\boldsymbol{y})
\end{equation}

\noindent where $\lambda$ is a linear interpolation along $n$ layers of NF: \(\lambda=k/n\), where $k = 1,\ldots, n$. \ Here \(u_{\boldsymbol{Z}}(\boldsymbol{y})\) denotes the prior energy for a standard Gaussian prior and \(u_{\boldsymbol{X}}(\boldsymbol{y})\) denotes the target energy computed using the error between the output of NF \(\boldsymbol{y}_{t,k}\) and the GRU predicted output \({\boldsymbol{y}}_{t(GRU)}\) at time $t$. 

\begin{equation}
    u_{\boldsymbol{Z}}(\boldsymbol{y}) = \frac{1}{2} \| \boldsymbol{y}_{t,k} \|^2 
\end{equation}

\begin{equation}
    u_{\boldsymbol{X}}(\boldsymbol{y}) = \left\| {\boldsymbol{y}}_{t(GRU)} - \boldsymbol{y}_{t,k} \right\|_2
\end{equation}

\noindent where, \(\mu_{\boldsymbol{X}}(\boldsymbol{y}) \propto \exp(-u_{\boldsymbol{X}}(\boldsymbol{y}))\) is an unnormalized energy-based probability distribution over $\boldsymbol{y}$.

\noindent In our proposal where GRU-NF is used for keypoint forecasting we use GRU predictions to define the target energy function. This is because GRUs provide temporally consistent, low-bias estimates that can effectively distinguish plausible keypoint trajectories making them well-suited for guiding our MCMC driven stochastic refinement.
\noindent The MCMC step is executed by first generating a sample from a proposal distribution \(q(. \mid .)\) such as a Gaussian centered at the current state: \(\boldsymbol{y}'_{t,k} \sim q(\boldsymbol{y}'_{t,k} \mid \boldsymbol{y}_{t,k})\) which is then accepted with the Metropolis-Hastings \cite{b24} acceptance probability based on the estimated potential function \(u_\lambda\):

\begin{equation}
    \alpha(\boldsymbol{y}_{t,k} \rightarrow \boldsymbol{y}'_{t,k}) = \min\left(1, \frac{\exp(-u_\lambda(\boldsymbol{y}'_{t,k}))}{\exp(-u_\lambda(\boldsymbol{y}_{t,k}))} \right)
\end{equation}

\noindent After running these steps $m=2$ times, the accepted samples are passed to the next NF layer in the sequence. The final sequence of transformations can be written as: 

\(\boldsymbol{z_t} \rightarrow \boldsymbol{y}_{t(NF)_{1}} \rightarrow \boldsymbol{y}_{t(MCMC)_{1}} \rightarrow \cdots \rightarrow \boldsymbol{y}_{t(NF)_{n}} \rightarrow \boldsymbol{y}_{t(MCMC)_{n}} \rightarrow \boldsymbol{y'_t}\).

\begin{figure}
\centering
    \includegraphics[width=0.5\textwidth,height=0.25\textheight]{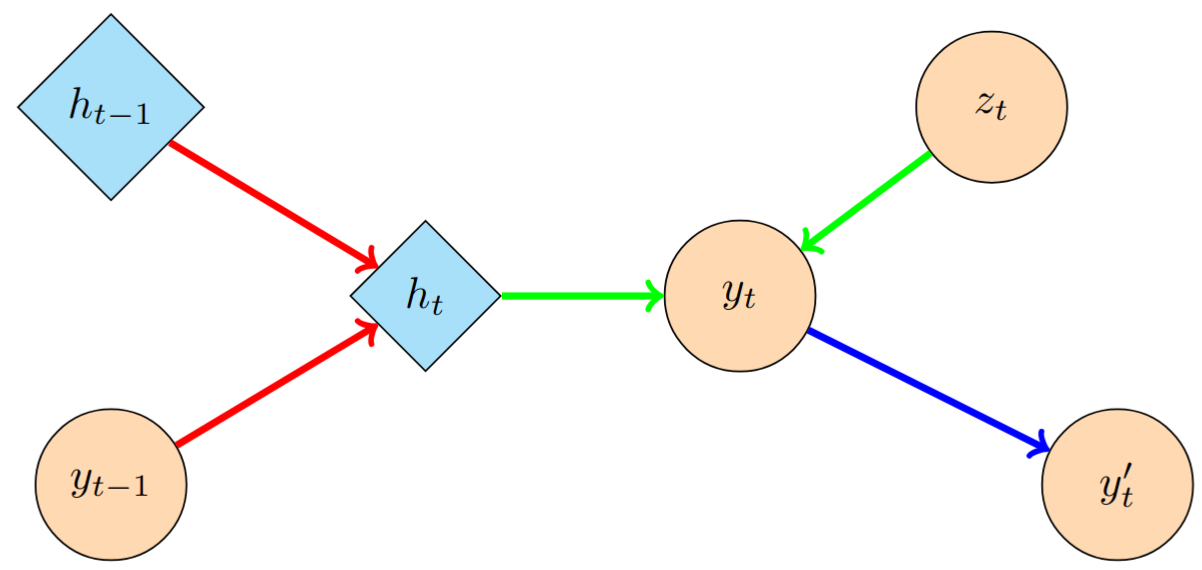}
    \caption{Processing steps in the proposed GRU-SNF framework showing the dependencies between the variables during inference. Red arrows show the GRU hidden state update, green arrows show the conditional sample generation with NF inverse transformation, and the blue arrow shows the MCMC refinement of NF output.}
    \label{fig:GRU-SNF}
\end{figure}

Figure \ref{fig:GRU-SNF} shows the processing steps in the proposed GRU-SNF framework.

\section{Results}

We evaluate the diversity-fidelity tradeoff of GRU-NF and GRU-SNF, using the VoxCeleb \cite{b25} and BAIR \cite{b26} datasets across multiple prediction horizons. For this purpose we train the baseline model GRU-NF on the VoxCeleb and BAIR datasets using 3883 videos and 5001 videos, and test them on 44 videos and 256 videos respectively. For both datasets the videos have 256 x 256 resolution. For each input sequence, we generate 100 samples during inference and assess performance in two stages. First, we compute energy distance between the low-dimensional keypoints extracted using the motion transfer pipeline of the First Order Motion Model (FOMM) \cite{b2} and the keypoints which are forecast with GRU-NF and GRU-SNF after they have been extracted using FOMM. Second, the diversity-fidelity tradeoff between the original videos and the generated videos using FOMM and prediction module is evaluated. Since longer prediction horizons contribute to greater bandwidth efficiency in real-time settings by reducing the frequency of keypoint transmissions, we evaluate the performance at different horizons. As the prediction horizon increases, the ambiguity in possible motions grows, requiring better multimodal modeling.

\subsection{Evaluation on predicted keypoints}

To quantify the distributional alignment between the predicted keypoint sequences $y, y^{'} \in Y\sim Q$ and ground truth keypoint sequences $x, x^{'} \in X\sim P$, we estimate the energy distance \cite{b27}:



\begin{equation}
\begin{split}
\mathcal{E}^2(P,Q)
&= 2\,\mathbb{E}_{\substack{x\sim P\\ y\sim Q}}\lVert x-y\rVert - \mathbb{E}_{x,x'\sim P}\lVert x-x'\rVert \\
 &\quad - \mathbb{E}_{y,y'\sim Q}\lVert y-y'\rVert
\end{split}
\end{equation}

\noindent The energy distance results are shown in Tables \ref{table:1} and \ref{table:2}. It is observed that across all prediction horizons and both datasets, GRU-SNF achieves consistently lower values than GRU-NF, demonstrating that even without retraining using stochastic layers, our framework generates more diverse and realistic keypoint sequences.

\begin{table}[htbp]
\centering
\caption{Energy distance results on predicted keypoints of VoxCeleb dataset}
\label{table:1}

\begin{tabular}
{|c||c|c|} 
 \hline
 \textbf{Prediction horizon (No. of input frames →}&\multicolumn{2}{|c|}{\textbf{Energy distance}} \\
\cline{2-3} 
\textbf{No. of output frames)} & \textbf{GRU-NF}& \textbf{GRU-SNF} \\
 \cline{2-3}
 \hline
 10 - 14  & 0.1917 & \textbf{0.1795} \\
 \hline
 8 - 16 & 0.3296 & \textbf{0.2872} \\ 
 \hline
 6 - 18 & 0.3924 & \textbf{0.3715} \\ 
 \hline
\end{tabular}
\vspace{0.2cm}
\end{table}

\begin{table}
\centering
\caption{Energy distance results on predicted keypoints of BAIR dataset}
\label{table:2}

\begin{tabular}
{|c||c|c|}
 \hline
 \textbf{Prediction horizon (No. of input frames →}&\multicolumn{2}{|c|}{\textbf{Energy distance}} \\
\cline{2-3} 
\textbf{No. of output frames)} & \textbf{GRU-NF}& \textbf{GRU-SNF} \\
 \cline{2-3}
 \hline
 7 - 8  & 8.232 & \textbf{7.3468} \\
 \hline
 6 - 9 & 14.9204 & \textbf{13.6054} \\ 
 \hline
 5 - 10 & 82.7789 & \textbf{80.9237} \\ 
 \hline
\end{tabular}
\vspace{0.2cm}
\end{table}

\subsection{Evaluation on generated videos}

Full video frames are very high-dimensional, making energy distance estimates unreliable owing to the curse of dimensionality. In this case to compare the model performances, we use the following two metrics:
\begin{itemize}
\item \textbf{APD (Average Pairwise Distance):} It measures diversity by computing the average pairwise distance across all predicted samples corresponding to a given video \cite{b6}. A model that generates more diverse samples attains a higher APD, indicating greater pairwise dissimilarity among its samples.
\item \textbf{MAE (Mean Absolute Error):} It measures video reconstruction fidelity versus ground truth by measuring the average of the absolute value of the errors across all pixels.
\end{itemize}
\noindent We keep C=20 samples among the generated D=100 samples per test video with the lowest MAE to examine the diversity–fidelity tradeoff (values of C,D can be adapted per application). We then compute the average MAE across these selected samples for each test video. For diversity, we compute APD as the mean pairwise distance among the same top-20 samples 
for each test video. Finally, we min–max normalize both MAE and APD using global minima and maxima computed over all test videos for both GRU-NF and GRU-SNF, placing the metrics on a common scale for direct comparison. Then we calculate the APD-to-MAE ratio using the standardized APD and MAE values for each test video and then average across all test videos, where higher values indicate better diversity without significant loss in fidelity. As shown in Tables \ref{table:3} and \ref{table:4}, the results across both the VoxCeleb and BAIR datasets indicate that GRU-SNF beats GRU-NF in 5 out of 6 cases and is therefore the superior framework. Additionally, GRU-SNF's relative performance over GRU-NF, measured by the APD-to-MAE ratio, improves with longer prediction horizons. In Table \ref{table:3}, it is observed that for the 10-14 horizon, GRU-SNF improves the APD-to-MAE ratio by 4.04\% over GRU-NF, for 8-16 the improvement is 24.02\% and for 6-18, the improvement rises to 36.90\%. We attribute this to the increasing prominence of multimodality in the high-dimensional video space, compared to the more constrained, lower-dimensional keypoint representations, where such effects are less pronounced.


\begin{table}
\centering
\caption{Diversity-fidelity tradeoff results on generated videos of VoxCeleb dataset}
\label{table:3}

\begin{tabular}
{|c||c|c|} 
 \hline
 \textbf{Prediction horizon (No. of input frames →}&\multicolumn{2}{|c|}{\textbf{APD-to-MAE ratio}} \\
\cline{2-3} 
\textbf{No. of output frames)} & \textbf{GRU-NF}& \textbf{GRU-SNF} \\
 \cline{2-3}
 \hline
 10 - 14 & 1.1811 & \textbf{1.2288} \\
 \hline
 8 - 16 & 1.0772 & \textbf{1.3359} \\ 
 \hline
 6 - 18 & 1.3106 & \textbf{1.7942} \\ 
 \hline
\end{tabular}
\vspace{0.2cm}
\end{table}

\begin{table}
\centering
\caption{Diversity-fidelity tradeoff results on generated videos of BAIR dataset}
\label{table:4}

\begin{tabular}
{|c||c|c|}
 \hline
 \textbf{Prediction horizon (No. of input frames →}&\multicolumn{2}{|c|}{\textbf{APD-to-MAE ratio}} \\
\cline{2-3} 
\textbf{No. of output frames)} & \textbf{GRU-NF}& \textbf{GRU-SNF} \\
 \cline{2-3} 
 \hline
 7 - 8  & 1.3238 & \textbf{1.3337} \\
 \hline
 6 - 9 & 1.0389 & \textbf{1.2076} \\ 
 \hline
 5 - 10 & \textbf{1.1591} & 1.0915 \\ 
 \hline
\end{tabular}
\vspace{0.2cm}
\end{table}

\begin{figure}[h]
\centering
\begin{minipage}{0.47\textwidth}
    \centering
    \includegraphics[width=\textwidth]{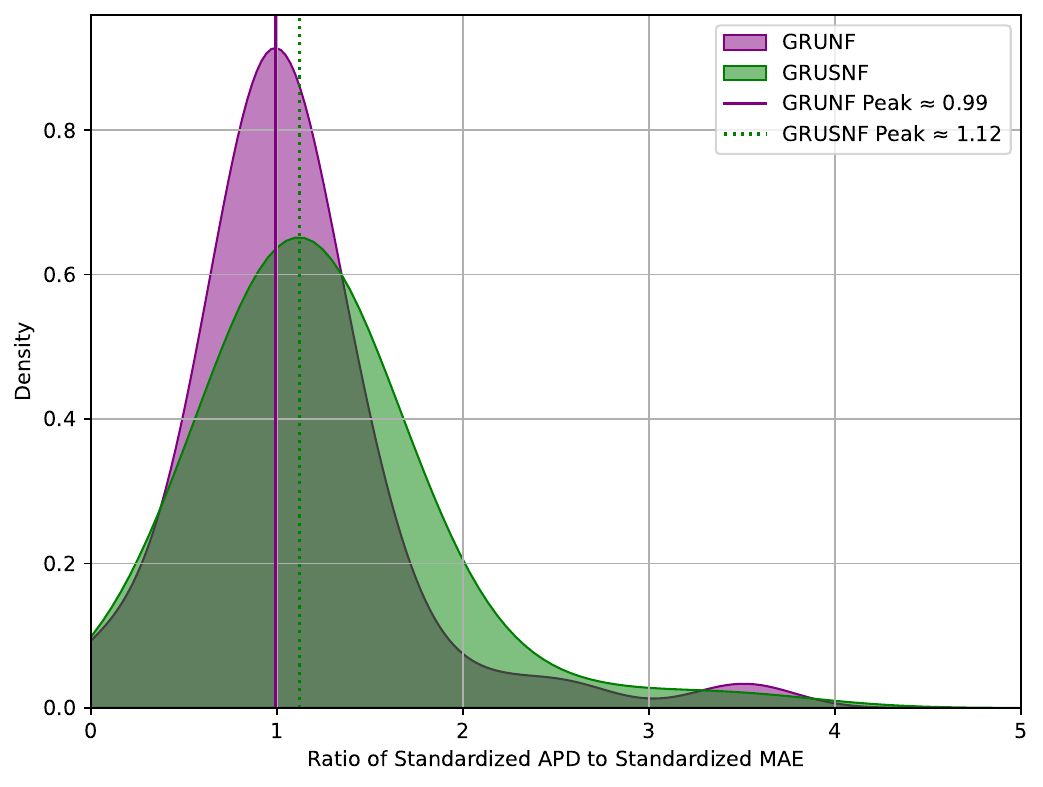}
    \caption{Comparison of APD to MAE ratio distributions for VoxCeleb dataset in 8--16 prediction horizon}
    \label{fig:APD_to_MAE_vox}
\end{minipage}%
\hfill
\begin{minipage}{0.47\textwidth}
    \centering
    \includegraphics[width=\textwidth]{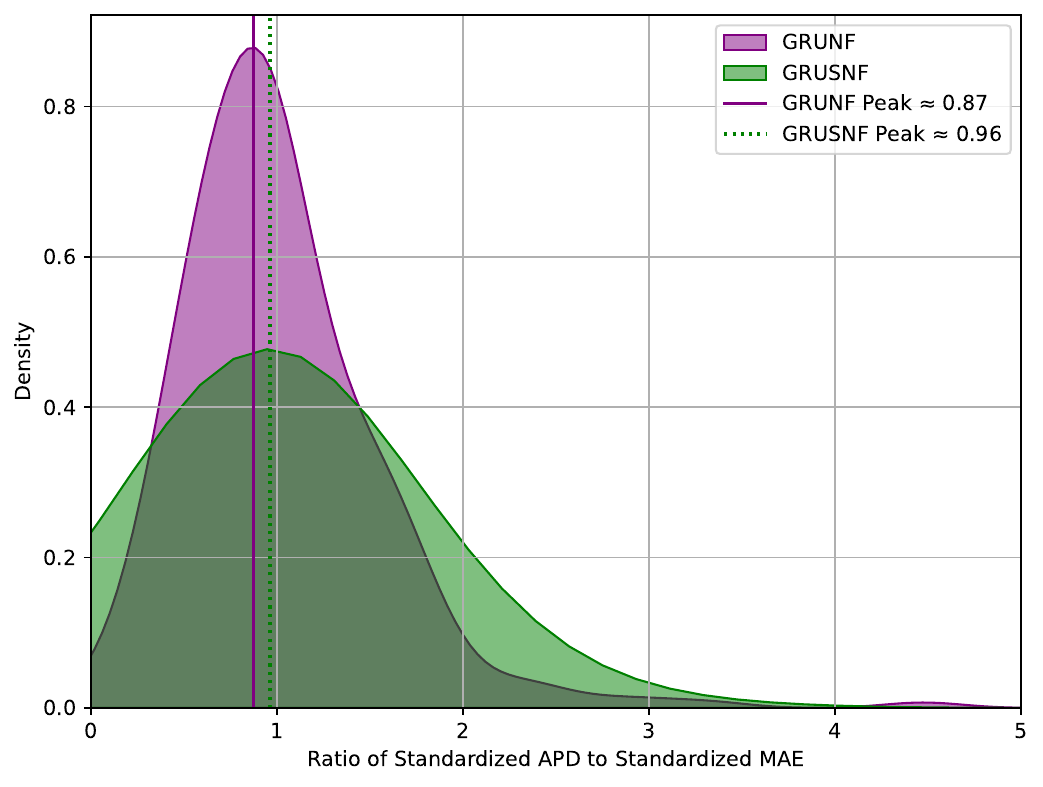}
    \caption{Comparison of APD to MAE ratio distributions for BAIR dataset in 6--9 prediction horizon}
    \label{fig:APD_to_MAE_bair}
\end{minipage}
\end{figure}

\noindent Figures \ref{fig:APD_to_MAE_vox} and \ref{fig:APD_to_MAE_bair} demonstrate the comparison of the APD to MAE ratio distributions for both GRU-NF and GRU-SNF using kernel density plots. Figure \ref{fig:APD_to_MAE_vox} and \ref{fig:APD_to_MAE_bair} show plots for VoxCeleb dataset in 8-16 prediction horizon and for BAIR dataset in 6-9 prediction horizon respectively. The plots imply that GRU-SNF provides a better tradeoff between diversity and fidelity since the density peak of GRU-SNF appears to the right of the peak of GRU-NF.

\begin{figure}[h]
    \centering
    \begin{minipage}{\textwidth}
        \centering
        \includegraphics[width=\textwidth, height=0.45\textheight]{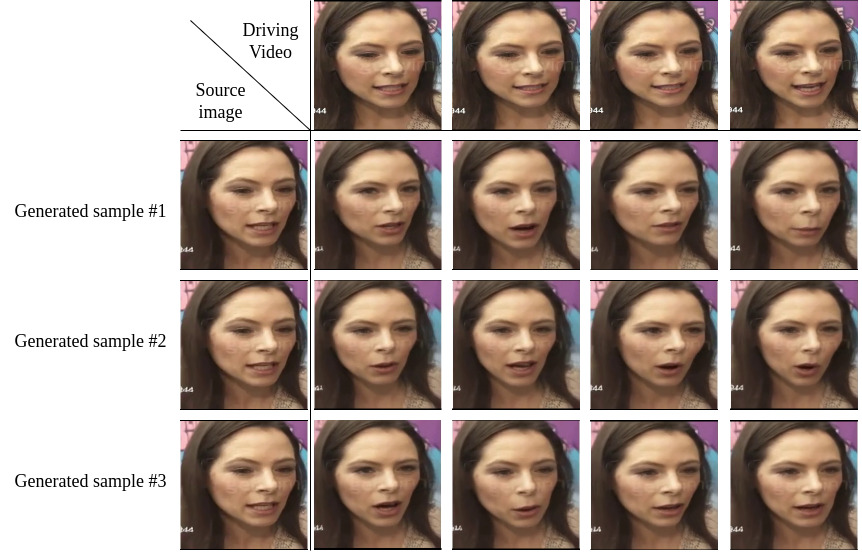}
    \end{minipage}
    
    \vspace{1.5em} 

    \begin{minipage}{\textwidth}
        \centering
        \includegraphics[width=\textwidth, height=0.45\textheight]{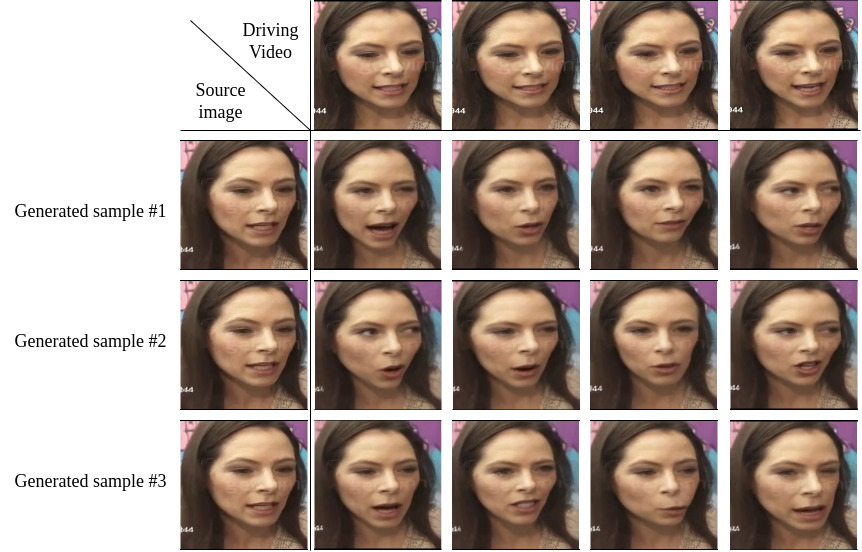}
    \end{minipage}
    
    \caption{Qualitative results for VoxCeleb dataset in generating diverse samples using GRU-NF (upper panel) and GRU-SNF (lower panel) for keypoint prediction on 8 - 16 prediction horizon. The first row represents the ground truth video and the second, third and fourth rows demonstrate the generated sample videos using the models.}
    \label{fig:diverse_prediction_recon}
\end{figure}

\noindent Figure \ref{fig:diverse_prediction_recon} shows the qualitative results of VoxCeleb dataset for prediction horizon 8-16. It is evident that the generated video frames using GRU-SNF show more diverse facial expressions compared to the frames generated using GRU-NF.

\section{Conclusion}
In this work we introduce GRU-SNF, which augments a trained GRU–NF with a limited number of inference-time MCMC refinements guided by a GRU-consistency based energy function. Our training-free method improves mode coverage and diversity–fidelity tradeoff, and demonstrates stronger gains at longer horizons. In video motion transfer, GRU-SNF boosts both keypoint and frame-level metrics while adding only small, tunable latency making it practical for real-time, low-bandwidth settings. Overall, our proposed inference-time stochastic refinement is a simple, general path to more useful multimodal forecasts in flow-based sequence models by overcoming flow expressivity limits without expensive retraining.

\section*{Acknowledgments}
Figure 1 was generated using ChatGPT and is for illustration only; it supports no data, results, or experimental claims. The authors would like to acknowledge the Pacific Research Platform, NSF Project ACI-1541349, and Larry Smarr (PI, Calit2 at UCSD) for providing the computing infrastructure.

\bibliographystyle{unsrt}  
\bibliography{references}

\end{document}